\documentclass{article} %
\usepackage[final]{colm2026_conference}

\usepackage{microtype}
\usepackage{hyperref}
\usepackage{url}
\usepackage{float}
\usepackage{diagbox}
\usepackage{pifont}
\usepackage{xcolor}
\usepackage{booktabs}
\usepackage{mdframed,xcolor}
\mdfdefinestyle{grayframe}{
  linecolor=lightgrayframe,
  linewidth=0.8pt,
  roundcorner=2pt,
  backgroundcolor=white
}

\usepackage{amsmath,amsfonts,bm}

\def\eqref#1{equation~\ref{#1}}

\def\1{\bm{1}}

\DeclareMathAlphabet{\mathsfit}{\encodingdefault}{\sfdefault}{m}{sl}
\SetMathAlphabet{\mathsfit}{bold}{\encodingdefault}{\sfdefault}{bx}{n}

\DeclareMathOperator*{\argmax}{arg\,max}

\definecolor{lightgray}{HTML}{808080}
\definecolor{perfup}{HTML}{008000}
\definecolor{perfdown}{HTML}{1d7b21}
\definecolor{myred}{HTML}{e60035}

\newcommand\redsout{\bgroup\markoverwith{\textcolor{red}{\rule[0.5ex]{2pt}{2pt}}}\ULon}
\newcommand\bluesin{\bgroup\markoverwith{\textcolor{green}{\rule[-0.5ex]{2pt}{2pt}}}\ULon}
\newcommand\redsin{\bgroup\markoverwith{\textcolor{red}{\rule[-0.5ex]{2pt}{2pt}}}\ULon}

\renewcommand{\sectionautorefname}{\S\kern-0.2em}
\renewcommand{\subsectionautorefname}{\S\kern-0.2em}
\renewcommand{\vec}[1]{\bm{#1}}

\newcommand{\textpr}[1]{\textcolor{violet!70!black}
{\texttt{\footnotesize #1}}}

\newcommand{\model}{M}
\newcommand{\simListener}{M_{judge}}
\newcommand{\image}{V}
\newcommand{\background}{B}
\newcommand{\information}{I}
\newcommand{\listener}{L}
\newcommand{\question}{Q}
\newcommand{\prompt}{P}
\newcommand{\des}{D}
\newcommand{\comment}{C}
\newcommand{\symbols}{\vec S}
\newcommand{\answer}{A}

\usepackage{inconsolata}
\usepackage{enumitem}
\usepackage{graphicx}
\usepackage{bbold}
\usepackage{dsfont}
\usepackage{caption}
\usepackage{subcaption}
\usepackage{xspace}
\usepackage{syntonly}
\usepackage{amsmath}
\usepackage{xcolor}
\usepackage{multirow} 
\usepackage{amssymb}
\usepackage{tabularx}
\usepackage{fvextra}
\usepackage[table]{xcolor}
\usepackage{colortbl}

\definecolor{baseblue}{RGB}{58,171,229}
\definecolor{pragmaticpurple}{RGB}{105,97,172}

\definecolor{zoey green}{rgb}{0.684,0.836,0.227}

\newcommand{\ignore}[1]{}

\usepackage{lineno}

\definecolor{darkblue}{rgb}{0, 0, 0.5}
\hypersetup{colorlinks=true, citecolor=darkblue, linkcolor=darkblue, urlcolor=darkblue}

\title{Pragmatics Meets Culture: Culturally-adapted Artwork \\
Description Generation and Evaluation}

\author{Lingjun Zhao, Dayeon Ki, Marine Carpuat \& Hal Daum\'e III \\
Department of Computer Science\\
University of Maryland\\
College Park, Maryland, USA \\
\texttt{\{lzhao123, dayeonki, marine, hal3\}@umd.edu} \\
}

\begin{document}

\ifcolmsubmission
\linenumbers
\fi

\maketitle

\begin{abstract}

Language models are known to exhibit various forms of cultural bias in decision-making tasks, yet much less is known about their degree of cultural familiarity in open-ended text generation tasks.
In this paper, we introduce the task of \textit{culturally-adapted art description generation}, where models describe artworks for audiences from different cultural groups who vary in their familiarity with the cultural symbols and narratives embedded in the artwork.
To evaluate cultural competence in this pragmatic generation task, we propose a framework based on culturally grounded question answering.
We find that base models are only marginally adequate for this task, but, through a pragmatic speaker model, we can improve simulated listener comprehension by up to 8.2\%.
A human study further confirms that the model with higher pragmatic competence is rated as more helpful for comprehension by 8.0\%.\footnote{Code and data will be released upon publication.}

\end{abstract}

\section{Introduction} \label{intro}

Culture can be characterized as a shared system of knowledge that shapes everyday reasoning and communication \citep{patterson2014making}. 
Effective cross-cultural communication relies on pragmatics,
which includes the ability to anticipate differences in shared knowledge and select utterances that minimize miscommunication \citep{grice1975logic, thomas2006cross, kecskes2014intercultural}. 
Similarly, when language models act as assistants, it is crucial that they communicate in a socially pragmatic manner, tailoring their messages to the listener’s perspective by accounting for prior knowledge, including cultural context \citep{ma2025pragmatics}.

While prior work has introduced inference-based tasks (e.g., natural language inference, multiple-choice question answering) to assess language models’ cultural knowledge \citep{yin2021broaden, huang2023culturally}, there remains a gap in introducing a text generation task that is appropriate  for a given listener’s cultural context to improve comprehension. 
Culturally-adapted text generation is challenging, particularly because the evaluation is difficult. 
Even when ground-truth annotations are available, reference-based evaluation metrics (e.g., BLEU \citep{papineni-etal-2002-bleu}) often fail to capture whether a model-generated text will be correctly interpreted by human listeners \citep{ma2025pragmatics, fried2023pragmatics}.

In this paper, we introduce a new task and evaluation framework for culturally-adapted text generation, and study whether language models can self-improve pragmatic competence.

\begin{figure}[t]
    \centering
    \includegraphics[width=\textwidth]{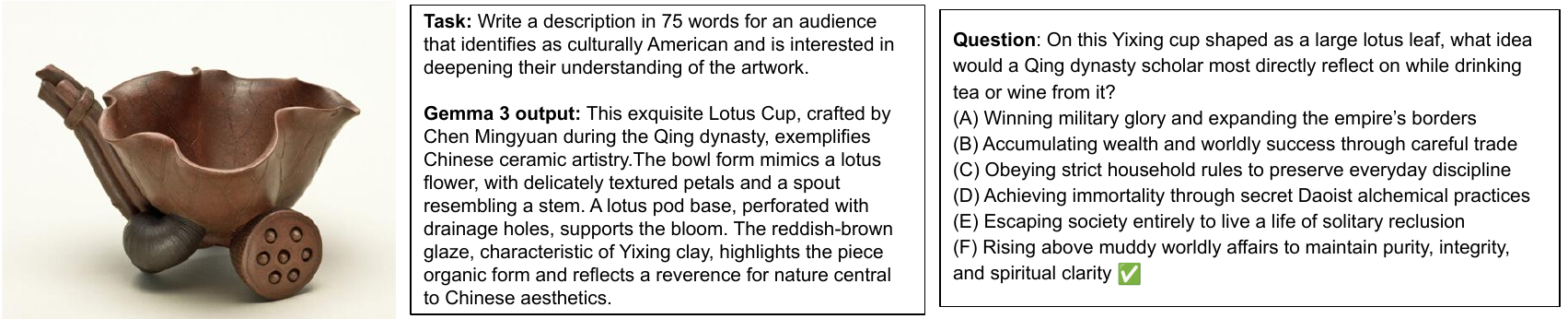} 
    \vspace{-0.5cm}
    \caption{
    Example of a Gemma3-generated description of artwork for audiences unfamiliar with the cultural context. The description fails to explain the symbolism of the lotus leaf and therefore lacks evidence to help them answer the culturally-attuned question.}
    \label{fig:teaser}
\end{figure}

First, we introduce a task for culturally-adapted text generation: generating descriptions of artwork tailored to different cultural audiences (\autoref{sec:task}), where we focus on American and Chinese cultural groups.
This task is suited for studying culturally-adapted text generation because visual artworks often encode rich cultural meanings\textemdash such as symbolism, historical narratives\textemdash that are rarely explicit but are readily interpreted by culturally-attuned audiences \citep{oparaocha2022, jain2017}.
As shown in \autoref{fig:teaser}, a visual-language model (VLM) fails to explain the cultural symbolism of the lotus leaf in a Chinese artwork, limiting its ability to help an American audience understand the piece. Success in this generation task therefore hinges on pragmatic competence: the ability to adapt to the listener's cultural common ground, bridge gaps in background knowledge, and facilitate comprehension.

Second, we propose a framework for evaluating the pragmatic competence of language models that supports both automatic and human evaluation (\autoref{sec:question_answer_eval}).
We hypothesize that the primary communicative goal of an description of artwork is to facilitate a listener’s understanding of the artwork.
To operationalize listener comprehension evaluation, we introduce an automated approach for constructing question–answering (QA) datasets grounded in the cultural symbols depicted in the artwork (\autoref{subsec:synthetic_eval_dataset}).
For automatic evaluation, we quantify pragmatic competence of VLMs by measuring the accuracy of an external simulated listener on the QA task when provided with each generated description.

Finding that VLMs exhibit limited pragmatic competence under automatic evaluation (\autoref{sec:exp_simulated_eval}), we investigate whether they can self-improve their pragmatic competence.
We propose a rational speech act–based approach \citep{goodman2016pragmatic} in which the same model serves as both speaker and simulated listener, with the speaker selecting descriptions that maximize simulated listener comprehension (\autoref{sec:theory_of_mind}).
Experimentally, we show that this self-improvement framework, when applied to Gemma3 \citep{gemma_2025}, LLaMA3.2 \citep{meta2024llama32vision}, and LLaVA‑OneVision \citep{li2025llavaonevision}, consistently outperforms the base models and improves external simulated listener comprehension in QA task by up to 8.2\% (\autoref{sec:exp_simulated_eval}).

Finally, recognizing that simulated listeners may not fully reflect human listeners, we design and conduct a human evaluation to examine how descriptions of artwork affect human comprehension (\autoref{sec:human_eval_design}, \autoref{sec:exp_human_eval}).
Our results show that participants rate descriptions from the model with higher pragmatic competence as more helpful for understanding the artwork than those generated by the base model, with an improvement of 8.0\%.
We also observe differences between simulated listener and human in comprehension, highlighting the need for better alignment between simulated listener evaluations and human comprehension.

\section{Related Work}

\noindent\textbf{Culturally aware language models.}
Prior work examines the capabilities of Artificial Intelligence (AI) tools for cross‑cultural understanding of human language \citep{saha2025reading, sarwari2024assessment}, while our focus is on adapting language models to audiences from different cultural backgrounds \citep{huang2023culturally, liu2025culturally}. 
Existing research analyzes cultural grounding in concepts \citep{liu2021visually, yin2021broaden, thapliyal2022crossmodal, li2024foodieqa, khanuja2024image, zhang-etal-2025-mllms} and values \citep{mohamed2022artelingo, mohamed2024no, tao2024cultural, ki2025multiple, yadav2025beyond, li2025culturellm}, but these efforts emphasize inference rather than generation. 
In contrast, while \citet{rooein2025biased} investigate cultural bias in generation, our goal is to bridge cultural knowledge gaps in generation.
\citet{shaikh2023modeling} introduce a culturally grounded reference game with a limited vocabulary and action space. Our work presents a text generation task with a richer vocabulary.
Culture is inherently multimodal\textemdash expressed through language, symbols, gestures, and artifacts \citep{liu2025culturally, pawar2025survey}; motivated by this, we introduce a task on culturally meaningful visual symbols and study how language models use such cues for culturally aware text generation.

\noindent\textbf{Pragmatics in grounded language generation.} 
Prior work in image captioning \citep{andreas2016reasoning, cohn2018pragmatically, tsvilodub2023evaluating, bao-etal-2022-learning, wang2021calibrate} and grounded instruction generation \citep{fried2018unified, dou2022foam, zhao2023cognitive} has primarily focused on contrastive reasoning\textemdash producing descriptions that distinguish a target image or trajectory from distractors\textemdash while assuming a uniform base of visual commonsense across audiences. 
However, pragmatic communication depends on audience-specific context, which can vary even when the visual input remains fixed. Our work addresses this gap by introducing a task and method that explicitly model audience differences and generate descriptions tailored to their distinct contextual expectations.

\noindent\textbf{Evaluating pragmatic language generation.}
Compared to question answering tasks that emphasize reasoning over context \citep{hu2023fine, qi-etal-2023-pragmaticqa, park2024multiprageval}, evaluating pragmatic language generation presents unique challenges. Standard reference-based metrics used in natural language generation often fail to capture whether the generated language will be correctly interpreted by human listeners \citep{ma2025pragmatics, fried2023pragmatics}, except in very simple cases\textemdash such as single-word outputs \citep{shaikh2023modeling, white2024communicate}.
To address this, researchers have modeled human responses using simulated users in domains like instruction generation and task-oriented dialogue \citep{lewis-etal-2017-deal, fried2018unified, zhao2023cognitive, kim2019codraw}. However, these evaluations are typically tied to specific datasets and task-specific metrics (e.g., scene reconstruction accuracy), making them difficult to generalize. 
Our work fills the gap of evaluating pragmatic success in language generation, by introducing a more general metric and automatic method for constructing evaluation datasets.

\noindent\textbf{Artwork in NLP.}
Art has long served as a challenging testbed for multimodal research, from early work on visual question-answering (VQA) \citep{guha-etal-2016-distorted, garcia_2018, garcia_2020} to recent efforts on cross-lingual \citep{ozaki-etal-2025-towards} or cross-cultural artwork explanation with VLMs \citep{zhang-etal-2025-creating, zhang-etal-2025-mllms}. Similarly, our work uses VLMs to generate descriptions of artwork, but focuses on culturally-adapted generation that explicitly aims to improve the pragmatic competence of these descriptions for diverse audiences.

\section{Culturally-adapted Text Generation Task and Evaluation}

In this paper, we introduce a culturally‑adapted text generation task to help bridge this gap and present a dataset (\autoref{sec:task}). We propose an evaluation framework (\autoref{sec:question_answer_eval}), and further develop a theory‑of‑mind–based modeling approach to improve generation (\autoref{sec:theory_of_mind}).
All relevant prompts are provided in \autoref{app:prompts}.

\subsection{Culturally-adapted Artwork Description Generation:  Task and Dataset}
\label{sec:task}

We are interested in generating descriptions of artwork that adapt to human listeners from different culture backgrounds, since visual artworks often embed rich cultural meanings-such as symbolism and historical narratives-that are not explicitly stated but resonate with
culturally-attuned audiences \citep{oparaocha2022, jain2017}.
Generating effective descriptions of artwork requires two key capabilities: interpreting implicit cultural cues and adapting them to the cultural backgrounds of diverse listeners. This makes artwork description generation a compelling task for studying culturally-adapted text generation. 
Success in this task depends on \textit{pragmatic competence}-the ability to adjust language to align with the common ground of a given cultural group.
This task challenges VLMs to move beyond the assumption of a uniform context across all audiences. Instead, it requires them to generate culturally-adapted descriptions that reflect the  common ground of specific cultural groups.

\paragraph{Task.}
We consider a description generation task for VLMs that can generate descriptions and answer questions.
An VLM $\model$ takes two components as \textit{input}: 
an artwork image $\image$, and a textual prompt $\prompt$.
The task is formulated by parameterizing $\prompt$ with (1)  basic information $\information$ about this artwork  e.g., \textpr{Title: \{TITLE\}, Artist: \{ARTIST\}, Date: \{DATE\}}  and (2) the cultural group $\listener$ of a human listener, e.g. \textpr{American} or \textpr{Chinese}.
The model then generates an artwork description as $\des$ to the human listener as \textit{output} with probability $\model(\des \mid \image, \prompt(\information, \listener))$.

For example, we format the description generation prompt $\prompt$ as: $\prompt (\information, \listener)$ = \textpr{ 
You will receive basic information about this artwork, and your task is to write an English description of the artwork in 75 words.
The description should be tailored for an audience that identifies as culturally \{$\listener$\} and is interested in deepening their understanding of the artwork.
Artwork information: \{$\information$\}.}

\paragraph{Dataset.}
\label{subsec:synthetic_eval_dataset} 
To generate the description, we start from a museum dataset with 6,399 artworks that contains basic metadata and a short comment $\comment$ for each visual artwork \footnote{The public domain dataset is from \url{https://www.artic.edu/open-access/open-access-images}.} We use the metadata to obtain  $\information$ for the description generation task.

We also use this dataset to construct a culturally-attuned question-answering dataset $\mathcal{D}_{\text{eval}}$ to enable evaluation of a listener's understanding of the artwork (\autoref{method:synthetic_dataset_creation}).
For each artwork, we first generate detailed background information $\background$ that emphasizes cultural aspects. To enrich $\background$ with embedded symbolism, we use GPT‑5 \citep{singh2025openaigpt5card} to identify culture symbols $\symbols = (S_1, \ldots, S_n)$ from the artwork, as symbolism often conveys rich cultural meaning \citep{oparaocha2022}. On average, 1.9 symbols are identified per artwork (e.g., the lotus leaf in \autoref{fig:teaser}). We then prompt GPT‑5 with $\symbols$ and the associated comment $\comment$ to generate the background, with an average length of 565 words.

We design questions in $\mathcal{D}_{\text{eval}}$ to probe the meanings of culture symbols $\symbols$ embedded in each artwork. Using GPT‑5, we follow the practices recommended in \citet{balepur-etal-2025-best} to generate multiple-choice question–answer triplets $(\question, \vec{\answer}, \hat{\answer})$ conditioned on the artwork image $\image$, generated background $\background$, and symbols $\symbols$. 
$\question$ is a textual question that requires understanding of the cultural meaning of an artwork, $\vec \answer = (\answer_1, \answer_2, ..., \answer_m)$ are a set of possible answers to $\question$, and $\hat{\answer}$ is the ground-truth answer to $\question$.
One triplet is generated per culture symbol, with six candidate answers of similar length that are randomly shuffled (an example in \autoref{fig:teaser}).
As a control, we construct a culturally agnostic dataset $\mathcal{D}_{\text{agnostic}}$ using only the artwork image $\image$, and we generate two triplets per artwork.

For evaluation, we use subsets of American and Chinese artworks, aligning with the cultural backgrounds of human participants. Specifically, we include 224 American artworks created after 1900 and 57 Chinese artworks created after 1400, as symbolism in earlier works may be unfamiliar even to same-culture participants. In total, $\mathcal{D}_{\text{eval}}$ contains 481 triplets, and $\mathcal{D}_{\text{agnostic}}$ contains 562 triplets.

\subsection{Evaluation Method for Pragmatic Description Generation}
\label{sec:question_answer_eval}

To evaluate whether a generated artwork description fulfills its pragmatic communicative goal—facilitating comprehension conditioned on the listener’s cultural background— 
we need an automatic evaluation method to support model development. 
This evaluation method shall not require ground-truth description annotation, as such annotation is difficult to obtain. Additionally, classical reference-based automatic natural language generation metrics such as BLEU are unable to measure whether or not generated language will be understood correctly by human listeners \citep{krahmer2010empirical, zhao2021evaluation}. 
Therefore, measuring the \textit{task success} of simulated listener using the description is a promising approach \citep{fried2023pragmatics}.

\paragraph{Question answering as evaluation.}
\label{method:synthetic_dataset_creation}
Previous work on measuring task success of pragmatic language generation tasks typically use human-annotated evaluation datasets, such as annotating target destination for grounded instruction generation, and target image for contrastive image captioning \citep{zhao2023define, andreas2016reasoning}. 
However, such annotations are difficult to obtain to measure artwork comprehension success, so we introduce a question-answering evaluation task to measure how well the model-generated description supports a listener's comprehension as task success.
We construct the question answering dataset $\mathcal{D}_{\text{eval}}$ using GPT-5, as described in \autoref{subsec:synthetic_eval_dataset}, an example is shown in \autoref{fig:teaser}.
This evaluation supports both simulated users and human users as listeners.

\paragraph{LLM as a judge.} 
For automatic evaluation of generated descriptions, we employ a simulated listener whose role is to answer questions about an artwork using the description, approximating how a human listener from a specific cultural group would respond. We operationalize this simulated listener with a vision–language model $\simListener$ that is \textit{different} from the description generation model $\model$.

A simulated listener $\simListener$ generates an entailment probability for each possible answer $\answer_i$, given an artwork image $\image$ and textual prompt $\prompt$ as input.
$\prompt$ is parameterized by: (1) a provided description $\des$, (2) a specific cultural group $\listener$ of the simulated listener, (3) question $\question$ about the artwork, and (4) answer $\answer_i$.
In this task, the output of simulated listener $\simListener$ is \textpr{Correct} or \textpr{Incorrect}.
The chosen answer is the one with the highest entailment probability:
\begin{equation}
\label{eq:basic_eval}
\answer^* = \argmax_{\answer_i} \; \simListener (\textpr{Correct} \mid V, \prompt(D, L, Q, \answer_i))
\end{equation}
To measure the task success, we compare the optimal answer $\answer^*$ chosen by simulated listener $\simListener$ with the ground-truth answer $\hat{\answer}$.

Our preliminary experiments show that the automatic evaluation method based on \autoref{eq:basic_eval} is not promising enough, as the model either ignores the description $\des$ or the listener's cultural group $\listener$.
To address this, 
we adopt a chain-of-thought–based evaluation strategy. For each candidate answer $\answer_i$, we provide two types of reasoning signals: (i) $R_L$, reasoning about whether the listener’s cultural group $\listener$ alone entails $\answer_i$ in the absence of the description, and (ii) $R_D$, reasoning about whether the description $\des$ entails $\answer_i$. Both $R_L$ and $R_D$ are generated by prompting simulated listener $\simListener$ (see \autoref{app:prompts} for details).
Under this formulation, the final answer is selected as:
\begin{equation}
\label{eq:cot_des_eval}
\answer^* = \argmax_{\answer_i} \; \simListener (\textpr{Correct} \mid V, \prompt(D, L, Q, \answer_i, R_D, R_L))
\end{equation}

However, when the simulated listener $\listener$ can already determine the correctness of $\answer_i$ confidently based solely on cultural knowledge, providing the description may introduce spurious or incorrect inferences.
To address this issue, when $\listener$ does not predict \textpr{unsure} for the answer entailment without description task (which also generates $R_L$), we evaluate answers without using the description for chain-of-thought prompting (see \autoref{app:prompts}).
\begin{equation}
\label{eq:cot_eval}
\answer^* = \argmax_{\answer_i} \; \simListener (\textpr{Correct} \mid V, \prompt(L, Q, \answer_i, R_L))
\end{equation}

\section{Pragmatic Description Generation with Theory-of-Mind Modeling}
\label{sec:theory_of_mind}
In this section, we describe our approach to generate pragmatic artwork description for the task in \autoref{sec:task}.
We first use role-playing persona prompting as our base speaker, and then use theory-of-mind modeling for pragmatic speaker, based on the rational speech act framework \citep{goodman2016pragmatic, fried2018unified, zhao2023define}.

\begin{figure}[t]
    \centering
    \includegraphics[width=\textwidth]{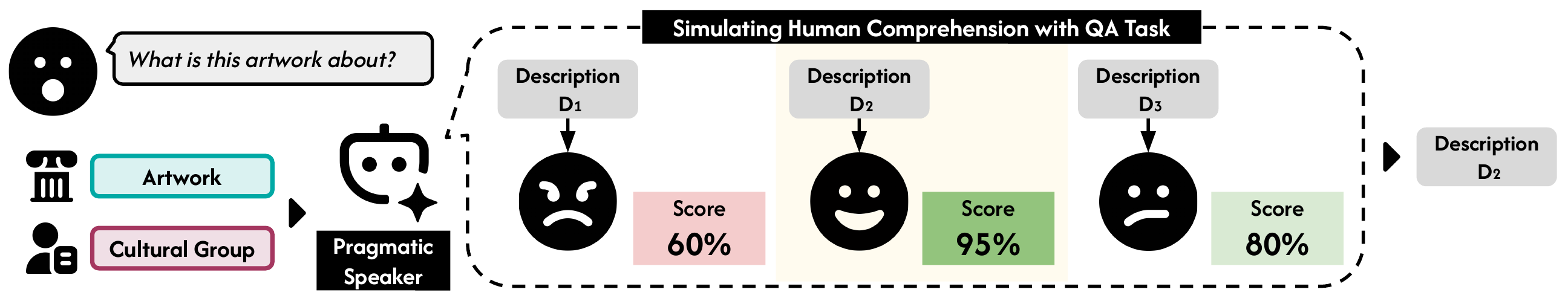} 
    \caption{
    Our approach uses a self-improving speaker model to generate pragmatic artwork descriptions. 
    Given an artwork and the listener's cultural group, the pragmatic speaker first samples multiple descriptions, then ranks them by simulating how the listener would answer culturally-attuned questions when provided with each description. The model then selects the description with the highest self‑evaluation score and presents it to the listener, either an external simulated listener or human listener.}
    \label{fig:model}
\end{figure}

\paragraph{\textcolor{baseblue}{\ding{108}} Base speaker.}
We use a pre-trained VLM $\model$ to generate a description $\des$ for the visual artwork $\image$ to a human listener from a specific cultural group $\listener$ with probability $\model(\des \mid \image, \prompt(\information, \listener))$, as described in \autoref{sec:task}. 
We use greedy decoding to generate description $\des$.

\paragraph{\textcolor{pragmaticpurple}{\ding{108}} Pragmatic speaker.}
When describing the artwork to human listener from cultural group $\listener$, the pragmatic speaker chooses a description $\des^*$ from the candidate description set that best enables $\listener$ to comprehend the artwork. This process is illustrated in \autoref{fig:model}.

To do this, we randomly sample $K$ descriptions from $\model$ using from the probability distribution $\model(\des \mid \image, \prompt(\information, \listener))$ to form the candidate description set $\vec \des = \{\des_1, ..., \des_K\} \cup \{\des\}$. 
Next, we use $\model$ again as simulated listener to answer questions related to the artwork, in order to approximate how human listener comprehend the artwork when provided a description $\des_i$:
\begin{equation}
\label{eq:pragmatic_speaker}
\des^* = \argmax_{\des_i \in \vec \des} \; \model (\textpr{Correct} \mid \image, \prompt(\des_i, \listener, \question, \hat{\answer}))
\end{equation}
\noindent where $\model$ generates a probability $\model (\textpr{Correct} \mid \image, \prompt(\des_i, \listener, \question, \hat{\answer}))$ that the  ground-truth answer $\hat{\answer}$ is entailed by the listener from a given cultural group $\listener$ for the question $\question$.
$\question$ and $\hat{\answer}$ are generated from \autoref{subsec:synthetic_eval_dataset}.
In this task, the output of simulated listener $\model$ is \textpr{Correct} or \textpr{Incorrect}.
The pragmatic speaker chooses description $\des^*$ as the one that has highest chance for the simulated listener to entail the correct answer. All prompts provided in \autoref{app:prompts}.

We also include the description generation likelihood to penalize potential hallucinations. Higher-likelihood descriptions also tend to be shorter, improving communication efficiency—an important pragmatic factor \citep{fried2023pragmatics} beyond simulated listener comprehension.
To achieve this, we design pragmatic speaker to choose description $\des^*$ as the one that maximizes a weighted product of probabilities under the description generation and simulated listener answer entailment:
\begin{equation}
\begin{aligned}
\des^* = \argmax_{\des_i \in \vec \des} \;& 
\model(\des_i \mid \image, \prompt(\information, \listener))^{\lambda} \\
&\times \model(\textpr{Correct} \mid \image, \prompt(\des_i, \listener, \question, \hat{\answer}))^{1-\lambda}
\end{aligned}
\label{eq:speaker_measure}
\end{equation}
\noindent where $\lambda \in [0, 1]$ is a hyperparamter tuned on development set.

\section{Experimental Setup}

\subsection{Models}
We evaluate three instruction-tuned VLMs for generating culturally-adapted artwork descriptions:
(i) Gemma3 12B \citep{gemma_2025}, (ii) LLaMA3.2-Vision 11B \citep{meta2024llama32vision}, and (iii) LLaVA-OneVision 7B \citep{li2025llavaonevision}. 
For the \textit{base speaker}, we use greedy decoding with a 75-word limit, truncating outputs exceeding this limit to the last complete sentence. 
For the \textit{pragmatic speaker}, we sample 10 candidate descriptions under the same word limit at temperature 1.0 for Gemma and LLaMA, and 20 descriptions with temperature 1.2 for LLaVA. 
We use Qwen2.5‑VL 7B \citep{wu2025qwen} as an external simulated user for evaluation.

\subsection{Human Evaluation Design}
\label{sec:human_eval_design}

Participants evaluate 12 artworks in stages: they first view the artwork alone and select any culture symbols they recognize, answering up to three questions about each (from \autoref{subsec:synthetic_eval_dataset}).
Next, they receive a description, sampled from either the base or pragmatic speaker model, and repeat the same tasks.
Finally, both descriptions are shown side by side, 
and we ask participants for preference ratings:
\begin{enumerate} [leftmargin=2em]
\item \textbf{Comprehension:} which description better helps you understand the main meaning of the artwork (and its symbolism)? 
\item \textbf{Knowledge gain:} which description provides more explanation of cultural symbols in the artwork that you did not already know?
\item \textbf{Prior knowledge:} which description provides more explanation of cultural symbols in the artwork that you were already familiar with?
\end{enumerate}

To aid understanding of culture-specific content in Chinese artworks, we provide English-to-Chinese translations for culture-specific terms identified via GPT-4o \citep{openai2024gpt4technicalreport} and resolved through Wikidata.
Additionally, to help participants quickly identify differences between the two descriptions, we highlight sentence pairs with similar meanings using SimCSE \citep{gao2021simcse} cosine similarity with threshold score of 0.8. Implementation details and survey screenshots are provided in \autoref{app:human_eval}.

We designed our survey on Qualtrics\footnote{\url{https://www.qualtrics.com/}} and recruited 10 participants from each of the two cultural groups (American and Chinese) via Prolific.\footnote{\url{https://www.prolific.com/}}
Each participant evaluates 12 artworks, including 2 quality-control tasks (explained in \S\ref{sec:exp_simulated_eval}), and was compensated at a rate of \$18/hour. Sessions failing below 75\% accuracy are discarded. In total, we retained 20 valid sessions (240 data points). All procedures were approved by our institution's IRB.

\section{Analysis}
We investigate the following questions:

\begin{enumerate}[leftmargin=20pt, itemsep=1pt]
    \item Can visual-language models (VLMs) self-improve their pragmatic competence?
    \item Do these improvements generalize to human judgments?
\end{enumerate}

\noindent To address Q1, we improve the model generated descriptions using theory-of-mind modeling, and evaluate them through an external simulated user study.
To answer Q2, we design and conduct a human evaluation.

\begin{table*}[]
\centering
\small
\begin{tabular}{lccccc}
\toprule
\textbf{QA Accuracy} 
& \multicolumn{2}{c}{\textbf{Culturally-agnostic questions}} 
& \multicolumn{2}{c}{\textbf{Culturally-attuned questions}} \\
\cmidrule(lr){2-3} \cmidrule(lr){4-5}
\diagbox{\textbf{Speaker Model}}{\textbf{Listener}}
& \textbf{Familiar} 
& \textbf{Unfamiliar} 
& \textbf{Familiar} 
& \textbf{Unfamiliar} \\
\midrule
No description  
 & \textcolor{gray}{89.9} & \textcolor{gray}{90.3} 
 & 90.2 & 71.5 \\
\midrule
Base Gemma \citep{gemma_2025}  
 & \textcolor{gray}{89.6} & \textcolor{gray}{89.9} 
 & \textcolor{gray}{88.6} 
 & \textcolor{baseblue}{73.2} \\
Pragmatic Gemma 
 & \textcolor{gray}{89.9} & \textcolor{gray}{90.1} 
 & \textcolor{gray}{\textbf{90.4}} 
 & \textcolor{pragmaticpurple}{\textbf{79.2}}$^{\dagger}$ \\
\midrule
Base LLaMA \citep{meta2024llama32vision} 
 & \textcolor{gray}{89.9} & \textcolor{gray}{90.6} 
 & \textcolor{gray}{87.5} 
 & \textcolor{baseblue}{70.9} \\
Pragmatic LLaMA 
 & \textcolor{gray}{90.1} & \textcolor{gray}{90.6} 
 & \textcolor{gray}{89.8$^{\dagger}$} 
 & \textcolor{pragmaticpurple}{74.0} \\
\midrule
Base LLaVA \citep{li2025llavaonevision}  
 & \textcolor{gray}{89.4} & \textcolor{gray}{89.9} 
 & \textcolor{gray}{87.1} 
 & \textcolor{baseblue}{64.4} \\
Pragmatic LLaVA  
 & \textcolor{gray}{89.1} & \textcolor{gray}{89.9} 
 & \textcolor{gray}{88.1} 
 & \textcolor{pragmaticpurple}{69.4}$^{\dagger}$ \\
\bottomrule
\end{tabular}
\caption{
Question-answering accuracy for culturally-agnostic and culturally-attuned artwork‑related questions, using descriptions generated by different speaker models. An external simulated listener (Qwen2.5‑VL), either familiar or unfamiliar with the artwork’s culture, answers the questions.
${\dagger}$ denotes when the pragmatic model has significantly higher accuracy than the base model ($p < 0.05$) as determined by a two-sample $t$-test. 
\textbf{Takeaway:} VLMs can self-improve their pragmatic competence when used as speaker models.
}
\label{tab:simulated_eval}
\end{table*}

\subsection{Can VLMs Self-improve Their Pragmatic Competence?}
\label{sec:exp_simulated_eval}

We employ a VLM different from the speaker model as an external simulated listener (user) to evaluate the speaker's pragmatic competence, providing each speaker generated description when answering questions about an artwork (\autoref{sec:question_answer_eval}).
See \autoref{sec:theory_of_mind} for the speaker models.
Results are reported in \autoref{tab:simulated_eval}, distinguishing between \textit{culturally-attuned}, which require specific cultural background knowledge to answer correctly, and \textit{culturally-agnostic} questions, which serve as a control condition and do not rely on cultural knowledge (see \autoref{subsec:synthetic_eval_dataset}). 
We partition results by the simulated listener's cultural familiarity with the artwork: the \textit{familiar} condition corresponds to users whose cultural group matches that of the artwork, while the \textit{unfamiliar} condition corresponds to users from a different cultural group.

\paragraph{Performance on control conditions.} 
Across all models and settings, QA accuracy on culturally-agnostic questions remains consistently high and similar between familiar and unfamiliar simulated listeners, regardless of whether a description is provided. 
This confirms that these questions are indeed culturally agnostic, as neither cultural group nor descriptions play a role in these control conditions.

For culturally-attuned questions in the familiar condition, there are often modest differences across rows. This is expected: listeners who already possess the relevant cultural knowledge can answer these questions reliably even without additional descriptions, limiting the potential benefits of generated descriptions.

\paragraph{VLMs can self-improve pragmatic competence.}
For culturally-attuned questions, unfamiliar listeners perform much worse than familiar listeners when no description is provided, reflecting the lack of requisite cultural knowledge. 
For the unfamiliar listeners, providing descriptions from the base Gemma speaker yields only minor improvements over the no description condition, suggesting these descriptions are insufficiently culturally-adapted to support comprehension.
By comparison, descriptions from by the pragmatic Gemma speaker lead to a substantial improvement in QA accuracy over the base model (73.2 to 79.2). 
This gain indicates that the pragmatic model produces more culturally-adapted descriptions with improved pragmatic competence, enabling listeners outside the culture to answer culturally-attuned questions more accurately. 
We observe a similar trend for the LLaMA.
For LLaVA, base model descriptions has lower performance relative to the no description condition, suggesting that suboptimal descriptions can hinder comprehension. The pragmatic LLaVA model mitigates this degradation, but overall performance remains bounded by the quality of the candidate description set.

\subsection{Do Simulation Improvements Generalize to Human Judgments?}
\label{sec:exp_human_eval}

\begin{figure}[]
    \centering
    \includegraphics[width=0.8\textwidth]{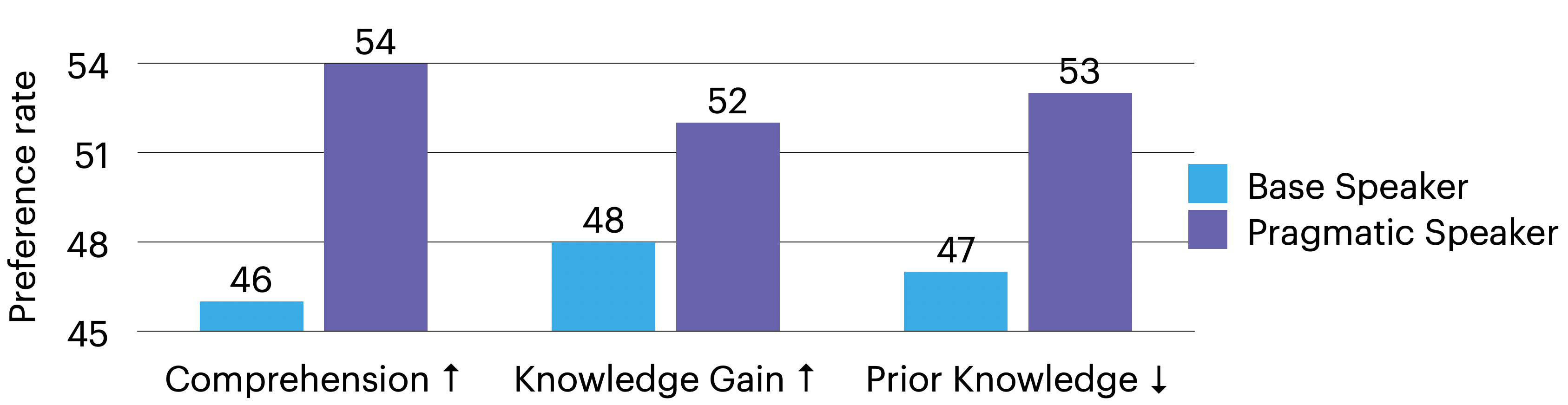} 
    \caption{
    Human users preference rates for descriptions generated by the base speaker and the pragmatic speaker. \textbf{Takeaway:} The pragmatic speaker improves user comprehension and introduces new information, but insufficiently accounts for users' existing knowledge. 
    }
    \label{fig:preference_chart}
\end{figure}

\begin{table*}[]
\centering
\small
\begin{tabular}{@{}llllc@{}}
\toprule
                                   & \textbf{Comprehension}      & \textbf{Knowledge Gain}     & \textbf{Prior Knowledge}    & \textbf{QA Accuracy Gain}   \\ \midrule
\textbf{Self-eval. Score}  & 0.08 (-0.06, 0.22) & 0.04 (-0.10, 0.18) & 0.06 (-0.08, 0.20) & 0.15 (0.05, 0.25)  \\
\bottomrule
\end{tabular}
\caption{Kendall's $\tau$ correlation between model self-evaluation score (\autoref{eq:speaker_measure}), human preference over a pair of descriptions, and human QA accuracy gain from using the description versus not using it. Ranges are 95\% confidence intervals.
\textbf{Takeaway:} Model self‑evaluation score has stronger correlation with human QA accuracy gain than with human preferences.}
\label{tab:correlation}
\end{table*}

We conduct human evaluation on descriptions generated by Gemma, as it achieves the highest accuracy on the culturally-attuned dataset in the simulated listener evaluation (\autoref{sec:exp_simulated_eval}). 

\noindent\textbf{Preference ratings.} 
As shown in  \autoref{fig:preference_chart}, participants prefer pragmatic speaker descriptions over base speaker descriptions in terms of improving their comprehension of the artwork, and providing more informative explanations of unfamiliar cultural symbols. 
This suggests the model with higher simulated pragmatic competence also tend to be more helpful for human understanding.
At the same time, descriptions from the pragmatic speaker model more often include explanations of cultural symbols that users already know, indicating redundancy. This points to a direction for future work: improving the pragmatic speaker's ability to tailor explanations more precisely to a human user's existing knowledge.

\begin{figure}[t]
\centering
\begin{subfigure}{\textwidth}
\includegraphics[width=\textwidth]
{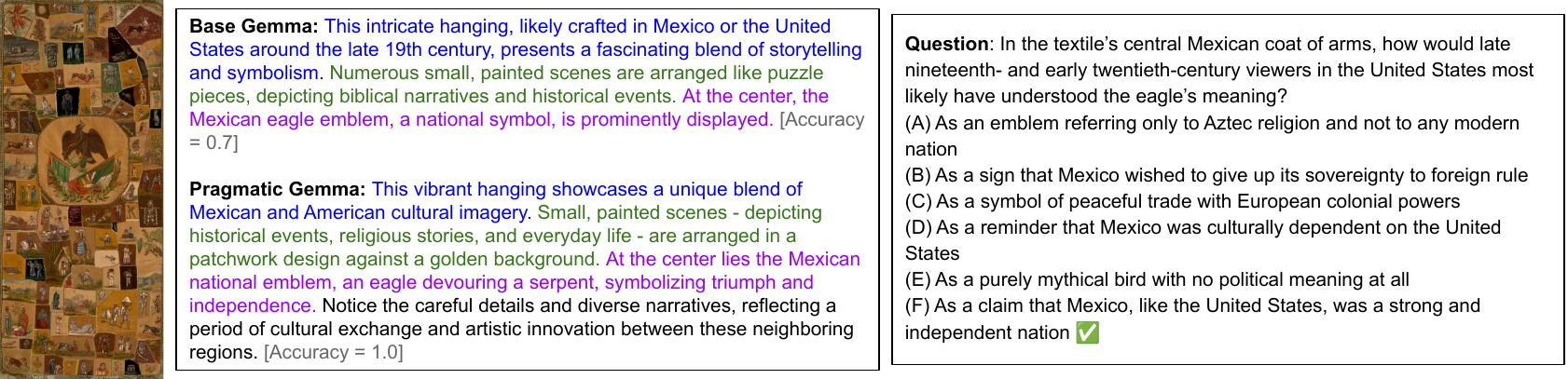} 
\centering
         \caption{The base model fails to generate descriptions that help culturally unfamiliar audiences answer the question, whereas the pragmatic model succeeds. Consistently, the simulated listener predicts that the base speaker's description is insufficient (choosing A), while the pragmatic model's description enables a correct response (choosing F).}
\end{subfigure}
\hfill \\
\begin{subfigure}{\textwidth}
\includegraphics[width=\textwidth]{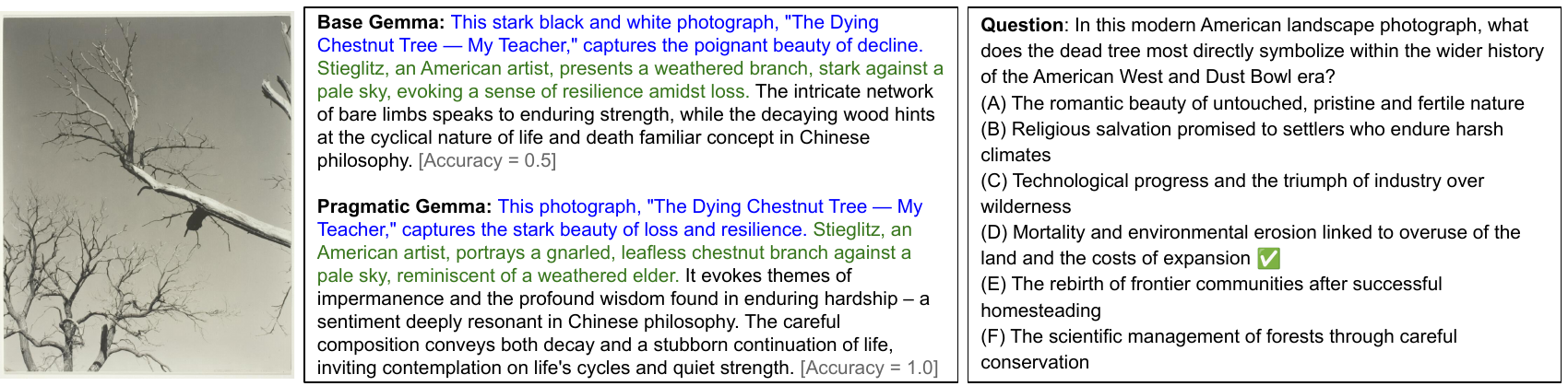}
\centering
         \caption{Both descriptions for culturally unfamiliar audiences lack sufficient evidence to support the correct answer. The simulator’s justification for option D using the pragmatic model's description is weak—``The introduction discusses themes of loss, resilience, and endurance, which align with the proposed answer.''—yet it still assigns D the highest probability (0.5; \autoref{eq:cot_des_eval}) among the options.}
\end{subfigure}
    \caption{The pragmatic speaker outperforms the base speaker in question-answering under an external simulated listener, both when (a) the listener aligns with human understanding and when (b) it does not. Colors denote semantically similar sentences.}
    \label{fig:qualitative_analysis}
\end{figure}

\noindent\textbf{Qualitative examples.}
\autoref{fig:qualitative_analysis} presents qualitative examples highlighting cases where the external simulated listener's interpretation of a description aligns with or diverges from that of a human user. 
It demonstrates difference between simulated and human understanding when descriptions are provided.
User QA accuracy is provided in \autoref{app:qa_accuracy}.

\noindent\textbf{Ablation: Correlation analysis.} 
Effective theory-of-mind modeling requires alignment between the model's simulated listener understanding and human comprehension. 
To assess this, we analyze correlations between the model's self-evaluated comprehension score (\autoref{eq:speaker_measure}) and human preference judgments across description pairs for the same artwork (\autoref{tab:correlation}).
We observe very weak correlations with human preferences, but stronger correlations with human QA accuracy gains, consistent with the self-evaluation metric being grounded in the QA task. 
This highlights the need for future research in developing simulated user metrics that better align with human judgments.

\section{Conclusion}

We introduce a new task for culturally-adapted text generation using artworks, as they often embed cultural meanings. 
We propose an evaluation framework for pragmatic competence, framing the communication goal as facilitating the listener’s understanding. 
We further present a self-improvement approach that enables language models to enhance their pragmatic competence, and demonstrate its effectiveness through simulated listener evaluations and human preference judgments.
Finally, we show that generated descriptions influence human comprehension differently than simulated user comprehension, underscoring the need for better alignment between simulated and human users.

\section*{Acknowledgments}

This material is supported by the NSF under Grant No. 2229885 (NSF Institute for Trustworthy AI in Law and Society, TRAILS).
We thank Yu Hou for suggestion on human evaluation interface, and Mark Riedl and Zichao Wang for discussion on pragmatics.

\bibliography{custom}
\bibliographystyle{colm2026_conference}

\clearpage
\appendix
\section{Appendix}
\subsection{Prompts}
\label{app:prompts}

\paragraph{Simulated Listener Entailment Task (\autoref{eq:basic_eval}).} We use the following entailment prompt for the simulated listener if the $\listener$ is the American cultural group, and use a Chinese translated version if $\listener$ is the Chinese cultural group:

\begin{Verbatim}[breaklines=true, breakanywhere=true, fontsize=\small, breaksymbol={}, formatcom=\color{violet!70!black}]
**Task:**
Imagine you are an {L}, also consider yourself culturally {L} and are {FAMILIARITY}.
Your job is to decide whether the proposed answer is 'Correct' or 'Incorrect' based on both the artwork and its introduction.
Others will read your evaluation, so ensure it reflects the perspective of the persona described above.

**Inputs:**
1. Question: {Q}
2. Introduction to the artwork: {D}
3. Proposed answer: {A_i}

**Assessments:**
The answer is (exactly one word: Correct or Incorrect; no additional text):
\end{Verbatim}

\noindent where the $\listener$ is a specified cultural group of the simulated listener, $\question$ is a textual question related to the artwork, $\des$ is a provided artwork description, $\answer_i$ is a possible answer to the question $\question$.

\paragraph{Simulated Listener Entailment Task using Chain-of-thought (\autoref{eq:cot_des_eval}).}
We first prompt the simulated user from culture $\listener$ to generate whether they will entail the answer and why:
\begin{Verbatim}[breaklines=true, breakanywhere=true, fontsize=\small, breaksymbol={}, formatcom=\color{violet!70!black}]
**Task:**
Imagine you are an {L}, also consider yourself culturally {L} and are {FAMILIARITY}.
Your job is to evaluate whether a proposed answer to a question about an artwork is correct based solely on what is visible in the artwork (without any external information):
- If you can confidently judge based on what is visible and your cultural knowledge, conclude correct or incorrect.
- If the artwork does not provide enough information or you lack the knowledge to decide, conclude unsure.
- Always mention the proposed answer directly in the reasoning, stating whether what is visible in the artwork supports, contradicts, or is insufficient to judge the proposed answer.
Others will read your evaluation, so ensure it reflects the perspective of the persona described above.

Return your answer as a single JSON string only, do not include ```json or any extra text. 
Output format (strict):
{{
  "Knowledge Check": "<one-sentence reasoning>",
  "Knowledge Check Final": "<correct / incorrect / unsure>",
}}

**Inputs:**
1. Question: {Q}
2. Proposed answer: {A_i}

**Outputs:**
\end{Verbatim}

Next, we prompt the model to generate whether description $\des$ entails the answer and why:

\begin{Verbatim}[breaklines=true, breakanywhere=true, fontsize=\small, breaksymbol={}, formatcom=\color{violet!70!black}]
**Task:**
You are given a question about an artwork, a proposed answer, and an introduction.
Provide assessment of the proposed answer: does the introduction explicitly support or contradict the proposed answer?
- If the introduction clearly confirms the proposed answer, conclude correct.
- If it clearly contradicts the proposed answer, conclude incorrect.
- If it does not explicitly confirm or contradict the proposed answer, conclude unsure.
- Always mention the proposed answer directly in the reasoning, stating whether the introduction supports, contradicts, or fails to address it.

Return your answer as a single JSON string only, do not include ```json or any extra text. 
Output format (strict):
{{
  "Information Check": "<one-sentence reasoning>",
  "Information Check Final": "<correct / incorrect / unsure>"
}}

**Inputs:**
1. Question: {Q}
2. Introduction: {D}
3. Proposed answer: {A_i}

**Outputs:**
\end{Verbatim}

We can then obtain the reasoning $R_L$ from ``Knowledge Check'', and $R_D$ from ``Information Check''. Finally, we prompt the model for answer entailment:

\begin{Verbatim}[breaklines=true, breakanywhere=true, fontsize=\small, breaksymbol={}, formatcom=\color{violet!70!black}]
**Task:**
Imagine you are an {L}, also consider yourself culturally {L} and are {FAMILIARITY}.
Your job is to decide whether the proposed answer is 'Correct' or 'Incorrect' based on both the artwork and its introduction.
- First, make your judgment by looking only at the artwork (without reading the introduction).
- If you cannot confidently decide based on the artwork alone, then refer to the introduction for further evaluation.
Others will read your evaluation, so ensure it reflects the perspective of the persona described above.

**Inputs:**
1. Question: {Q}
2. Introduction to the artwork: {D}
3. Proposed answer: {A_i}

**Assessments:**
1. Artwork-only judgment from your perspective: {R_L}
2. Judgment from the introduction: {R_D}
Therefore, the answer is:
\end{Verbatim}

For all of the above prompts we use a Chinese translated version if $\listener$ is the Chinese cultural group.

\paragraph{Simulated Listener Entailment Task using Chain-of-thought without Description (\autoref{eq:cot_eval}).}

\begin{Verbatim}[breaklines=true, breakanywhere=true, fontsize=\small, breaksymbol={}, formatcom=\color{violet!70!black}]
**Task:**
Imagine you are an {L}, also consider yourself culturally {L} and are {FAMILIARITY}.
Your job is to decide whether the proposed answer is 'Correct' or 'Incorrect'.
Others will read your evaluation, so ensure it reflects the perspective of the persona described above.

**Inputs:**
1. Question: {Q}
2. Proposed answer: {A_i}

**Assessments:**
{R_L}
Therefore, the answer is:
\end{Verbatim}

We use a Chinese translated prompt version if $\listener$ is the Chinese cultural group.

\paragraph{Simulated Listener Answer entailment for Pragmatic Speaker (\autoref{eq:pragmatic_speaker}).}
We use the following prompt for generate the answer using a visual-language model as simulated listener:

\noindent \textpr{
**Task:**\\
Imagine you are an \{\listener\}, also consider yourself culturally \{\listener\}.
Your job is to decide whether the proposed answer is 'Correct' or 'Incorrect' based on both the artwork and its introduction.
Others will read your evaluation, so ensure it reflects the perspective of the persona described above.\\\\
**Inputs:**\\
1. Question: \{\question\}\\
2. Introduction to the artwork: \{\des\}\\
3. Proposed answer: \{$\hat{\answer}$\}\\\\
**Assessments:**\\
The answer is (exactly one word: Correct or Incorrect; no additional text):
}

\noindent where the $\listener$ is a specified cultural group of the simulated listener, the $\question$ is a question related to the artwork, $\des$ is a model-generated description, and $\hat{\answer}$ is the ground-truth answer for $\question$.
We use a Chinese translated prompt version if $\listener$ is the Chinese cultural group.

\paragraph{Culture Symbol Generation (\autoref{subsec:synthetic_eval_dataset}).}
We use the following prompt for obtaining culture symbols using GPT-5.1:

\noindent \textpr{
**Task:**\\
Given an artwork and a target culture, analyze the visible elements and classify them into two categories:\\
1. Symbols: Visual elements in the artwork that are widely recognized symbols within \{school\} culture.\\
2. Non-symbols: Visual elements in this artwork that are not cultural symbols within \{school\} culture.\\\\
**Instructions:**\\
- Use concise, specific names for elements.\\
- Remove duplicates and merge close variants.\\
- For each symbolic element, provide a short, plain-language meaning.\\
- If no symbolic elements are present, return an empty object for "symbols".\\
- The "non-symbols" field must always contain at least one element.\\
- Do not include explanations, commentary, or formatting outside the JSON.}

\noindent where the \textpr{school} in the prompt is the cultural group of the corresponding artist.

\paragraph{Artwork Background Information Generation (\autoref{subsec:synthetic_eval_dataset}).} We use the following prompt for generating background information using GPT-5.1:

\noindent \textpr{
**Task:** 
You will be given an art comment and a list of cultural symbols related to this artwork. 
Based on this, generate a comprehensive cultural and historical background for the artwork.
Your response should:\\
- Include information about key figures, significant events, locations depicted.\\
- Explain the meaning of each provided cultural symbol within the context of \{school\} culture. \\
Do not discuss any symbols that are not included in the provided cultural symbols list.\\
- Write a self-contained background without referencing the original comment or the list cultural symbols directly.\\\\
**Art Comment:**\\
\{\comment\}\\\\
**Cultural Symbols:**\\
\{$\symbols$\}\\\\
**Background:**\\
}
\noindent where the \textpr{school} in the prompt is the cultural group of the corresponding artist, $\comment$ is the art comment provided in the dataset, and $\symbols$ is the culture symbols obtained from the visual artwork.

\paragraph{Question Answering Evaluation Dataset Generation (\autoref{subsec:synthetic_eval_dataset}).}
We use the following prompt for generate the culturally-attuned question answering dataset $\mathcal{D}_{\text{eval}}$ using GPT-5.1:

\begin{Verbatim}[breaklines=true, breakanywhere=true, fontsize=\small, breaksymbol={}, formatcom=\color{violet!70!black}]
**TASK:**
You will receive background information and a list of cultural symbols related to this artwork. 
Based on this information, create challenging, fact-based multiple-choice questions that focus on the cultural symbols in the artwork. 
Follow these guidelines:
1. Question:
    - Create one question per cultural symbol in the provided list.
    - Each question should require understanding of both the artwork and the provided background information.
    - The questions must be self-contained, without referencing terms like "background information".
    - The questions should be independent of each other. If no cultural symbols are provided, return an empty list.
2. Answer choices:
    - Provide 1 correct answer and 5 plausible but incorrect answers.
    - All the answers should be similar in length, but clearly distinct in meaning.
    - Avoid answers that are too vague, overly specific, opinion-based, or misleading.
3. Language:
    - Keep the language clear and the vocabulary simple.

Return your output as a list of JSON objects in the following strict format, do not include ```json or any extra text:
[
    {{
        "subject": "the symbolic element in the artwork",
        "question": "question text",
        "answer": "the correct answer",
        "type": "symbolism",
        "plausible_answers": [
            "plausible answer 1",
            "plausible answer 2",
            "plausible answer 3",
            "plausible answer 4",
            "plausible answer 5"
        ]
    }},
]

Example:
**Background:**
This Japanese woodblock print, likely from the Edo period (1603--1868), depicts a tranquil winter landscape where travelers make their way through snow-covered countryside. The Edo era was a time of significant cultural growth in Japan, marked by the rise of the merchant class and the widespread popularity of \textit{ukiyo-e}, an art genre that portrayed scenes of daily life, nature, and travel. In the print, the figures wear traditional clothing such as straw hats and winter robes, while a horse carrying goods suggests a commercial trip or pilgrimage, both common during that time. Horses in Japanese folklore often symbolize travel, endurance, and at times divine guidance. The subdued tones of black and gray, paired with selective touches of color on the horse, travelers, signs, and seals, create a contemplative atmosphere and draw attention to important narrative elements. Snow in Japanese art frequently represents purity and calm, reflecting both Shinto and Buddhist aesthetics. The red seals and \textit{kanji} inscriptions serve artistic and documentary functions, indicating authenticity and possibly identifying the artist or location. Overall, the print embodies the Edo period's appreciation for nature, seasonal transitions, and the quiet dignity of everyday life.
**Cultural Symbols:**
horse, snow
**Output:**
[
    {{
        "subject": "Horse",
        "question": "In the Edo-period winter scene, what does the horse most strongly symbolize?",
        "answer": "Travel and endurance, and sometimes divine guidance",
        "type": "symbolism",
        "plausible_answers": [
            "Wealth and prosperity in rural communities",
            "Strength and dominance in feudal society",
            "Agricultural abundance during winter months",
            "Protection against harsh seasonal conditions",
            "Status and power of the merchant class"
        ]
    }},
    {{
        "subject": "Snow",
        "question": "In Japanese art, what does snow typically represent in a scene like this?",
        "answer": "Purity and tranquility, reflecting Shinto and Buddhist aesthetics.",
        "type": "symbolism",
        "plausible_answers": [
            "Isolation and hardship as common themes in winter travel",
            "Impermanence and beauty as central ideas in seasonal change",
            "Economic decline and scarcity as effects of cold weather",
            "Celebration and joy as part of traditional winter festivals",
            "Divine punishment and suffering as warnings in folklore"
        ]
    }}
]

**Background:**
{B}
**Cultural Symbols:**
{S}
**Output:**
\end{Verbatim}

\noindent where the $\background$ is the background information of the artwork, and $\symbols$ are the culture symbols embedded in the artwork.

\paragraph{Culturally-agnostic Question Answering Dataset Generation (\autoref{subsec:synthetic_eval_dataset}).}
We use the following prompt for generate the culturally-agnostic question answering dataset $\mathcal{D}_{\text{agnostic}}$ using GPT-5.1:

\begin{Verbatim}[breaklines=true, breakanywhere=true, fontsize=\small, breaksymbol={}, formatcom=\color{violet!70!black}]
TASK: Based on the artwork, generate two easy fact-based multiple-choice questions that meet the following criteria:
- Each question must be answerable solely by observing the artwork, without requiring any external context such as cultural or historical background.
- Base each question on important content from the artwork; avoid trivial content.
- Each question should be independent—avoid overlapping content between the two.
- Each question should have one correct answer and five plausible but incorrect answer choices. 
- All the answer choices should be similar in length but clearly distinct in meaning.
- Avoid answers that are too vague, overly specific, opinion-based, or misleading.
- Keep the language clear and the vocabulary simple.

Return your output as a list of JSON objects in the following strict format:
[
    {
        "question": "question text",
        "answer": "a short sentence as the answer",
        "type": "easy",
        "plausible_answers": [
            "plausible answer 1",
            "plausible answer 2",
            "plausible answer 3",
            "plausible answer 4",
            "plausible answer 5"
        ]
    },
]

**Example Output:**
[
    {
        "question": "What type of weather is shown in the artwork?",
        "answer": "Snow is falling and the ground is covered in snow",
        "type": "easy",
        "plausible_answers": [
            "Rain is falling and the ground is wet",
            "The sun is shining and the sky is clear",
            "Strong winds are blowing through the trees",
            "Fog is covering the landscape and houses",
            "Lightning is striking near the mountain"
        ]
    },
    {
        "question": "What are the people in the artwork doing?",
        "answer": "Walking along a snowy path with a horse carrying goods",
        "type": "easy",
        "plausible_answers": [
            "Fishing in a river near the snowy trees",
            "Dancing in a field covered with snow",
            "Building a house beside the snowy road",
            "Sitting under a tree and reading a book",
            "Climbing a mountain with heavy backpacks"
        ]
    }
]

**Output:**
\end{Verbatim}

\paragraph{Culture-specific Item Identification (\autoref{sec:human_eval_design}).}
We use the following prompt for detecting any culture-specific terms in both base and pragmatic descriptions using GPT-4o:

\noindent \textpr{
**Task:**\\
From the Chinese artwork description below, identify words that are only culturally specific to Chinese whose translation into Chinese would help non-Chinese audiences better understand the artwork. \\ \\
**Description:* \{description\} \\ \\
**List of culturally specific word:*}

\subsection{Human Users Performance in Question-answering}
\label{app:qa_accuracy}
\begin{table*}[t]
\centering
\small
\begin{tabular}{lccccc}
\toprule
\textbf{QA Accuracy} & \multicolumn{2}{c}{\textbf{Familiar}} & \multicolumn{2}{c}{\textbf{Unfamiliar}} \\
\cmidrule(lr){2-3} \cmidrule(lr){4-5}
\textbf{Speaker} & \textbf{W.o. Description } & \textbf{W. Description } & \textbf{W.o. Description } & \textbf{W. Description } \\
\midrule
Base Gemma   & 35.9 (24.0–47.7) & 39.9 (27.7–52.1) & 29.8 (18.1–41.6) & 36.4 (23.4–49.4) \\
Pragmatic Gemma & 47.6 (37.3–57.9)   & 47.0 (35.5–58.5)  & 34.5 (24.1–45.0) & 40.0 (28.6–51.4) \\
\bottomrule
\end{tabular}
\caption{Human user QA accuracy, ranges are 95\% confidence intervals.
\textbf{Takeaway:} Users perform better on culturally familiar artworks than on unfamiliar ones, consistent with simulated evaluation. Descriptions help understanding of culturally unfamiliar artworks, while differences between models are not  significant.}
\label{tab:qa_accuracy}
\end{table*}

As shown in \autoref{tab:qa_accuracy}, when no artwork descriptions are provided, users achieve lower QA accuracy on artworks from unfamiliar cultures compared to artworks from cultures with which they are familiar. This finding is consistent with the external simulated evaluation reported in \autoref{tab:simulated_eval}.
We additionally find that descriptions improve user QA accuracy for culturally unfamiliar artworks. However, unlike the simulated evaluation, the performance differences between models are not  significant.
\autoref{fig:qualitative_analysis} presents qualitative examples highlighting cases where the external simulated listener’s interpretation of a description either aligns with or diverges from that of a human listener. 
It demonstrates difference between simulated and human understanding when descriptions are provided.

\subsection{Human Evaluation Details}
\label{app:human_eval}

Screenshots of the human evaluation survey, ordered to match the presentation flow shown to participants, are provided in \autoref{fig:human_eval}.

\paragraph{Cultural-Specific Term Translation.}
To aid participants' understanding of culture-specific content in Chinese artworks, we provide English-to-Chinese translations for culture-specific terms identified in the descriptions using GPT-4o (avg. 7.42 terms for base descriptions, 7.76 for pragmatic descriptions). Each identified term is translated via Wikidata by querying the term, taking the top-1 match, and retrieving its Chinese label.

\paragraph{Sentence Mapping.}
To help participants quickly identify differences between the two descriptions, we use SimCSE \citep{gao2021simcse} sentence embeddings to compute pairwise cosine similarity across sentences from both descriptions, highlighting sentence pairs with the highest similarity score above 0.8 as semantically similar. 

\begin{figure}[p]
    \centering
    \includegraphics[width=0.8\textwidth]{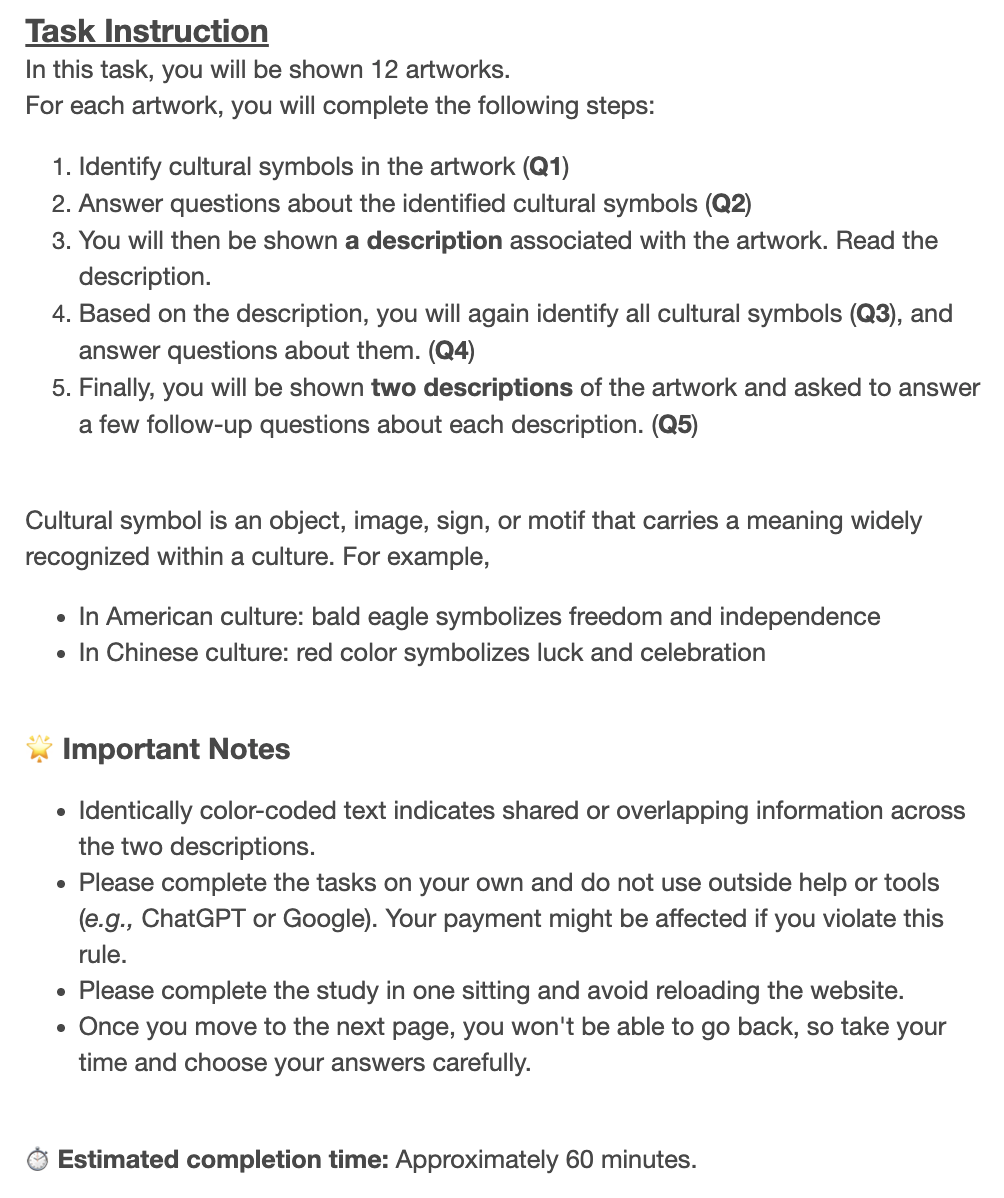}
    \caption{Screenshot of the human evaluation pipeline. Each participant first reads the task instructions and important notes.}
    \label{fig:human_eval}
\end{figure}

\begin{figure}[p]
    \ContinuedFloat
    \centering
    \includegraphics[width=0.7\textwidth]{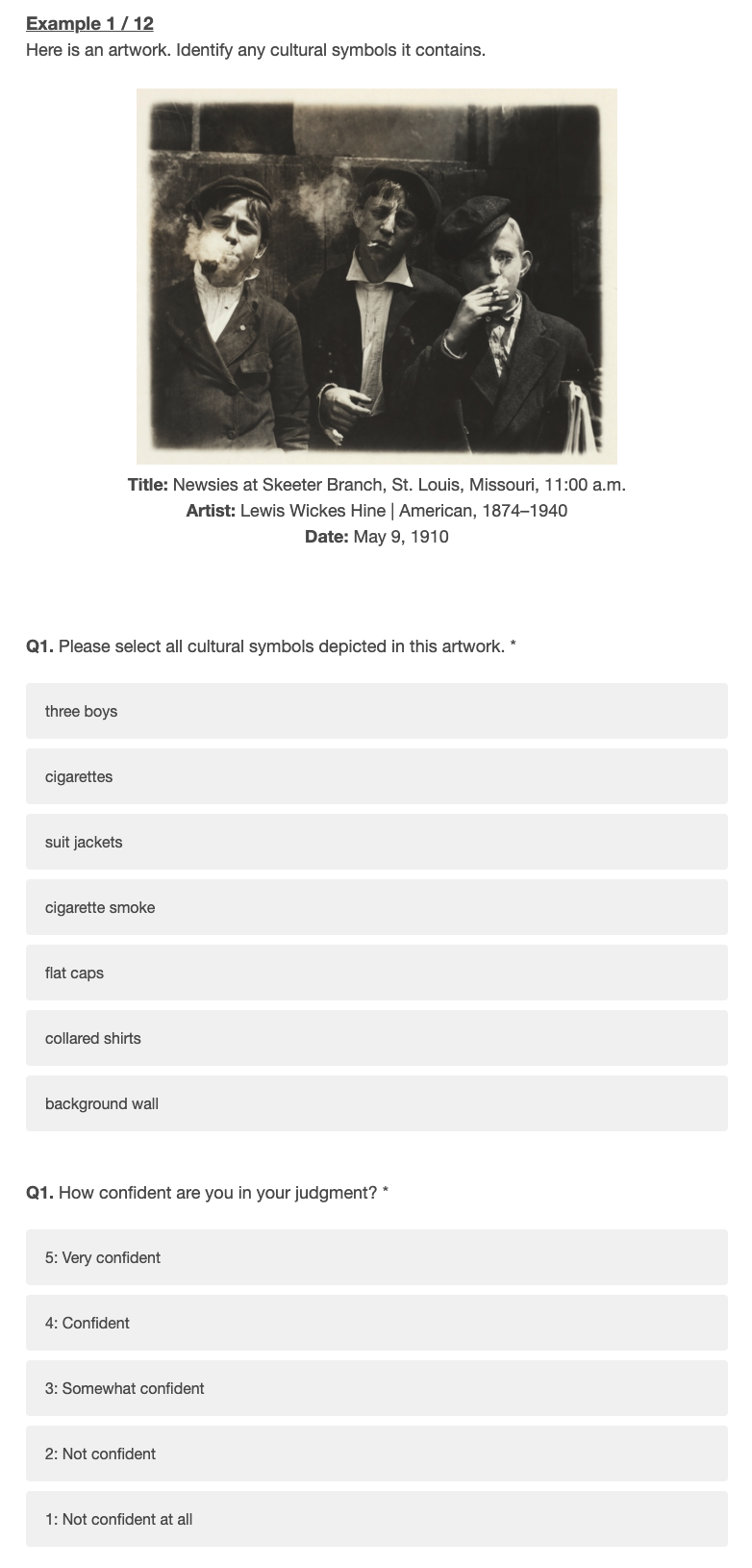}
    \caption{Participants select culture symbols from a list of visual elements, without any descriptions.}
\end{figure}

\begin{figure}[p]
    \ContinuedFloat
    \centering
    \includegraphics[width=0.7\textwidth]{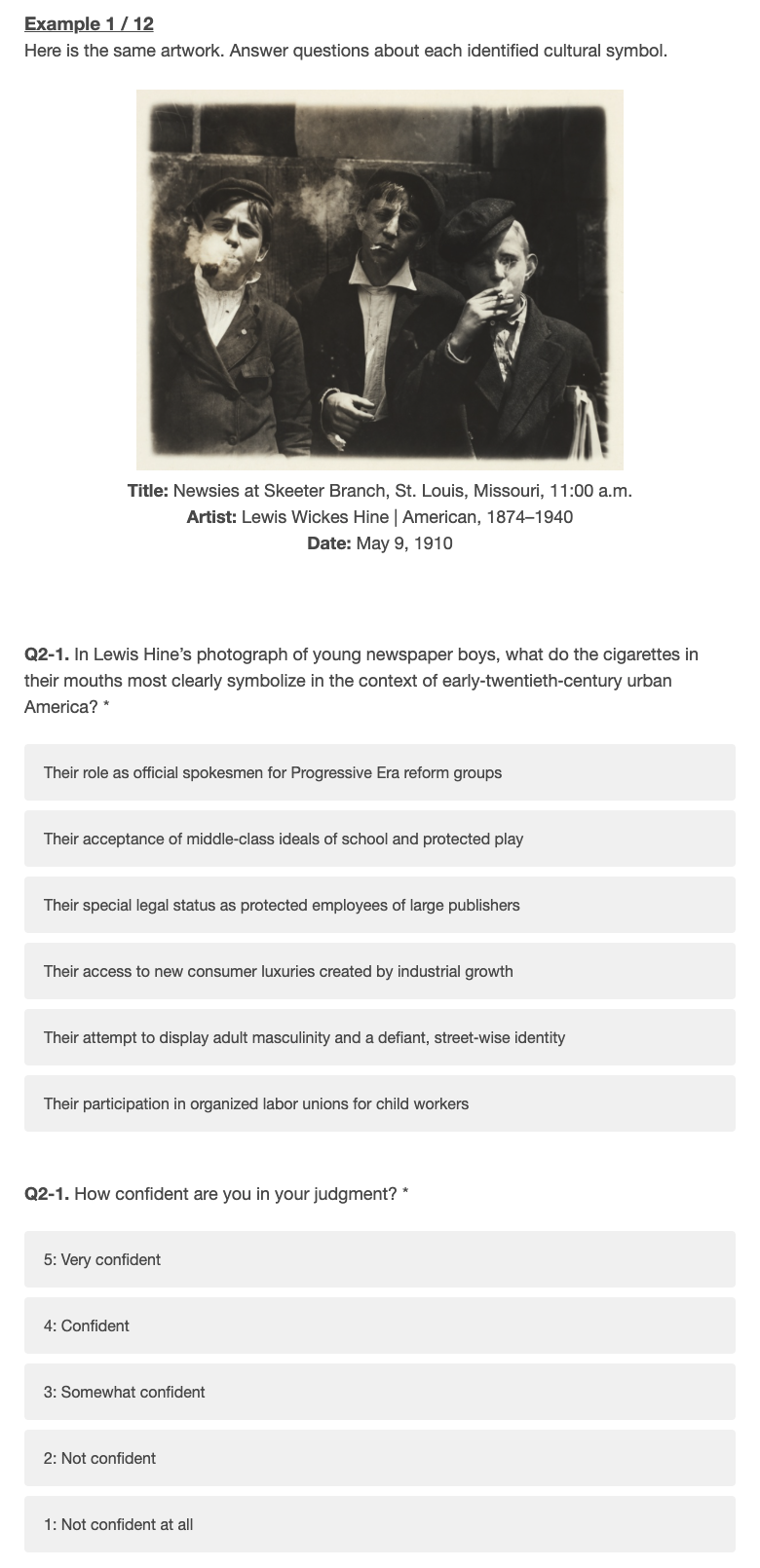}
    \caption{Participants answer three questions about each symbol, without being provided any artwork description.}
\end{figure}

\begin{figure}[p]
    \ContinuedFloat
    \centering
    \includegraphics[width=0.7\textwidth]{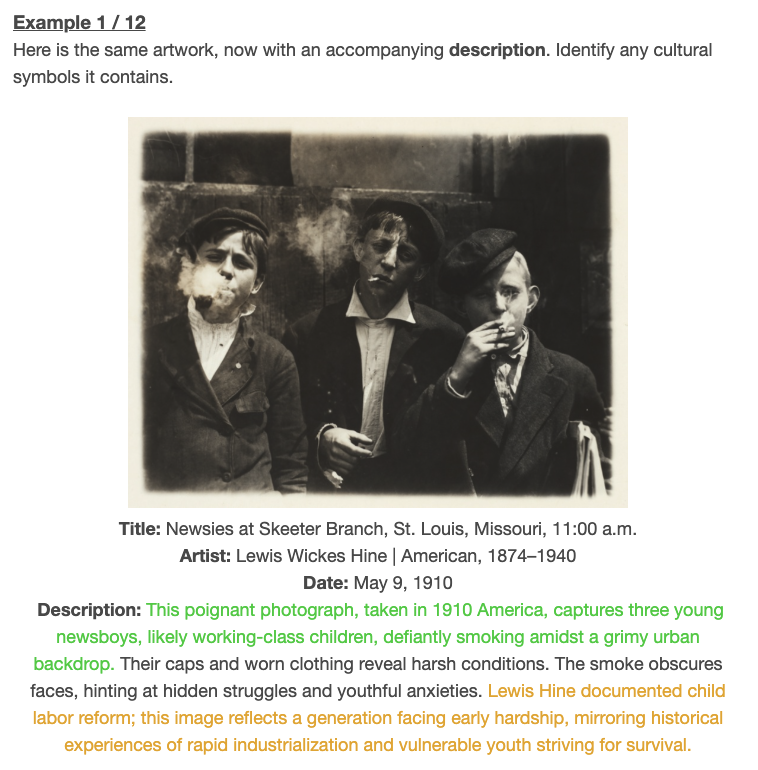}
    \caption{Participants are then shown a randomly sampled artwork description, either from the base or the pragmatic speaker model, and asked to complete the same tasks.}
\end{figure}

\begin{figure}[p]
    \ContinuedFloat
    \centering
    \includegraphics[width=0.8\textwidth]{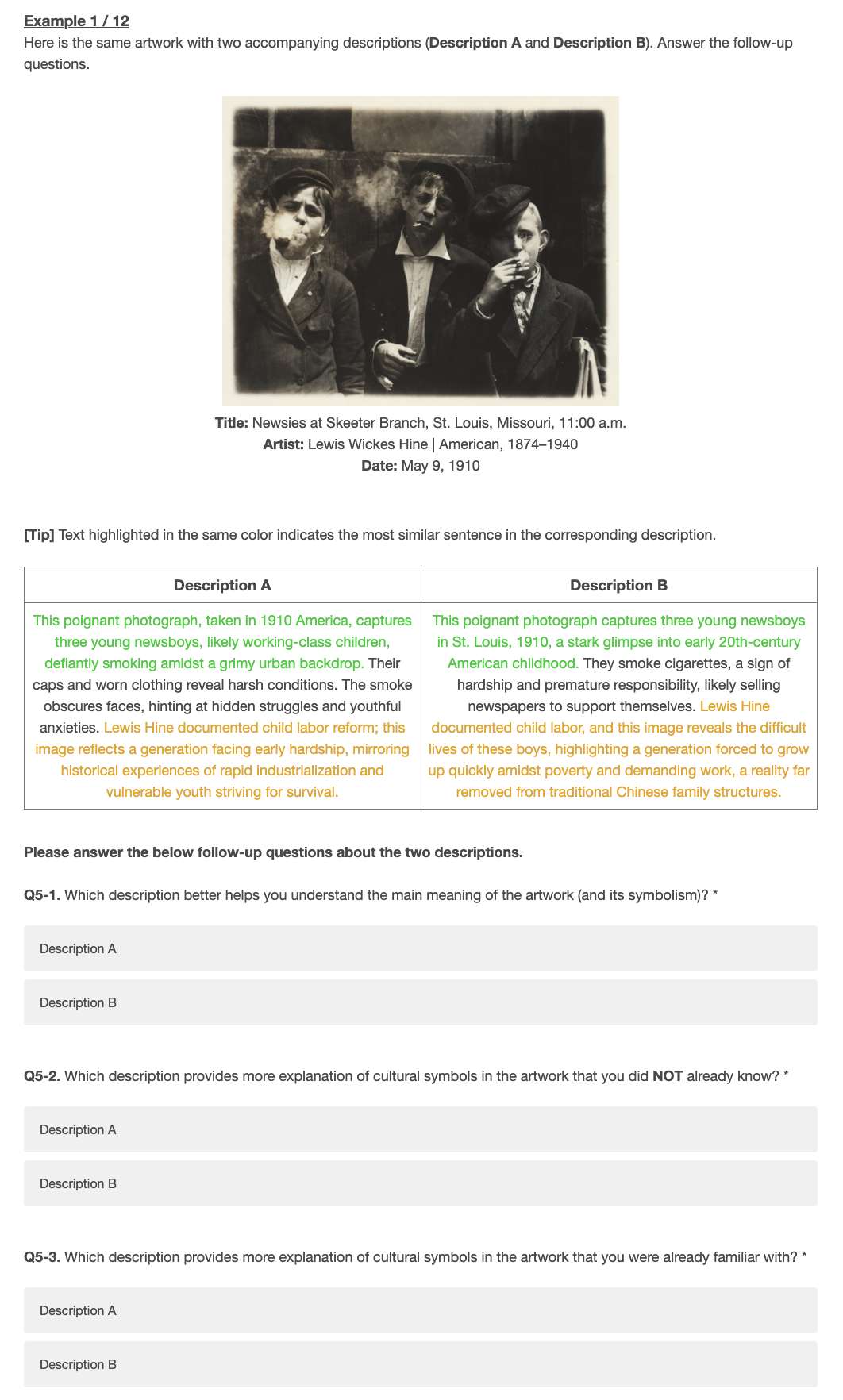}
    \caption{Participants make subjective judgments given two descriptions generated by the base and pragmatic models.}
\end{figure}

\end{document}